\documentclass[letterpaper]{article} 
\usepackage{aaai2026}  
\usepackage{times}  
\usepackage{helvet}  
\usepackage{courier}  
\usepackage[hyphens]{url}  
\usepackage{graphicx} 
\usepackage{natbib}  
\usepackage{caption} 
\usepackage{algorithm}
\usepackage{algorithmic}
\usepackage{amsfonts}
\usepackage{amsmath}
\usepackage{multirow}
\usepackage{booktabs}

\usepackage{newfloat}
\usepackage{listings}
\newcommand{\bfu}{\mathbf{u}}
\newcommand{\bfv}{\mathbf{v}}
\newcommand{\bfx}{\mathbf{x}}
\newcommand{\bfy}{\mathbf{y}}
\newcommand{\bfh}{\mathbf{h}}
\DeclareCaptionStyle{ruled}{labelfont=normalfont,labelsep=colon,strut=off} 
\floatstyle{ruled}
\newfloat{listing}{tb}{lst}{}
\floatname{listing}{Listing}
\title{Angular Gradient Sign Method: \\Uncovering Vulnerabilities in Hyperbolic Networks}
\author{
  Minsoo Jo\textsuperscript{\rm 1},
  Dongyoon Yang\textsuperscript{\rm 2}\footnotemark[1],
  Taesup Kim\textsuperscript{\rm 1}\thanks{Corresponding Authors.}
}
\affiliations{
  \textsuperscript{\rm 1}Graduate School of Data Science, Seoul National University\\
  \textsuperscript{\rm 2}AI Advanced Technology, SK hynix
}

\begin{document}

\maketitle

\begin{abstract}
Adversarial examples in neural networks have been extensively studied in Euclidean geometry, but recent advances in \textit{hyperbolic networks} call for a reevaluation of attack strategies in non-Euclidean geometries. Existing methods such as FGSM and PGD apply perturbations without regard to the underlying hyperbolic structure, potentially leading to inefficient or geometrically inconsistent attacks.
In this work, we propose a novel adversarial attack that explicitly leverages the geometric properties of hyperbolic space. Specifically, we compute the gradient of the loss function in the tangent space of hyperbolic space, decompose it into a radial (depth) component and an angular (semantic) component, and apply perturbation derived solely from the angular direction.
Our method generates adversarial examples by focusing perturbations in semantically sensitive directions encoded in angular movement within the hyperbolic geometry. Empirical results on image classification, cross-modal retrieval tasks and network architectures demonstrate that our attack achieves higher fooling rates than conventional adversarial attacks, while producing high-impact perturbations with deeper insights into vulnerabilities of hyperbolic embeddings. This work highlights the importance of geometry-aware adversarial strategies in curved representation spaces and provides a principled framework for attacking hierarchical embeddings. 
\end{abstract}
\section{Introduction}
Deep neural networks have achieved remarkable success across a wide range of domains.
However, they are also known to be highly sensitive to adversarial examples~\cite{Szegedy, fgsm}. 
These are specially crafted inputs (i.e., images) that include small, intentional perturbations designed to fool the model into making incorrect predictions.
Despite their effect on the model, such perturbations are often imperceptible to human observers, making them appear visually or semantically identical to the original data.
This vulnerability has motivated the development of numerous attack methods such as FGSM~\cite{fgsm} and PGD~\cite{pgd}, which generate input perturbations by leveraging gradients of the loss function.
While these methods apply perturbations to the input, they fundamentally rely on the assumption that the model’s representation space is Euclidean. 
As a result, they compute perturbation directions using gradients defined in a space with zero curvature (i.e., Euclidean geometry), which may not accurately reflect the underlying non-Euclidean structure of more complex models.

\begin{figure}[t]
\centering
\includegraphics[width=0.68\columnwidth]{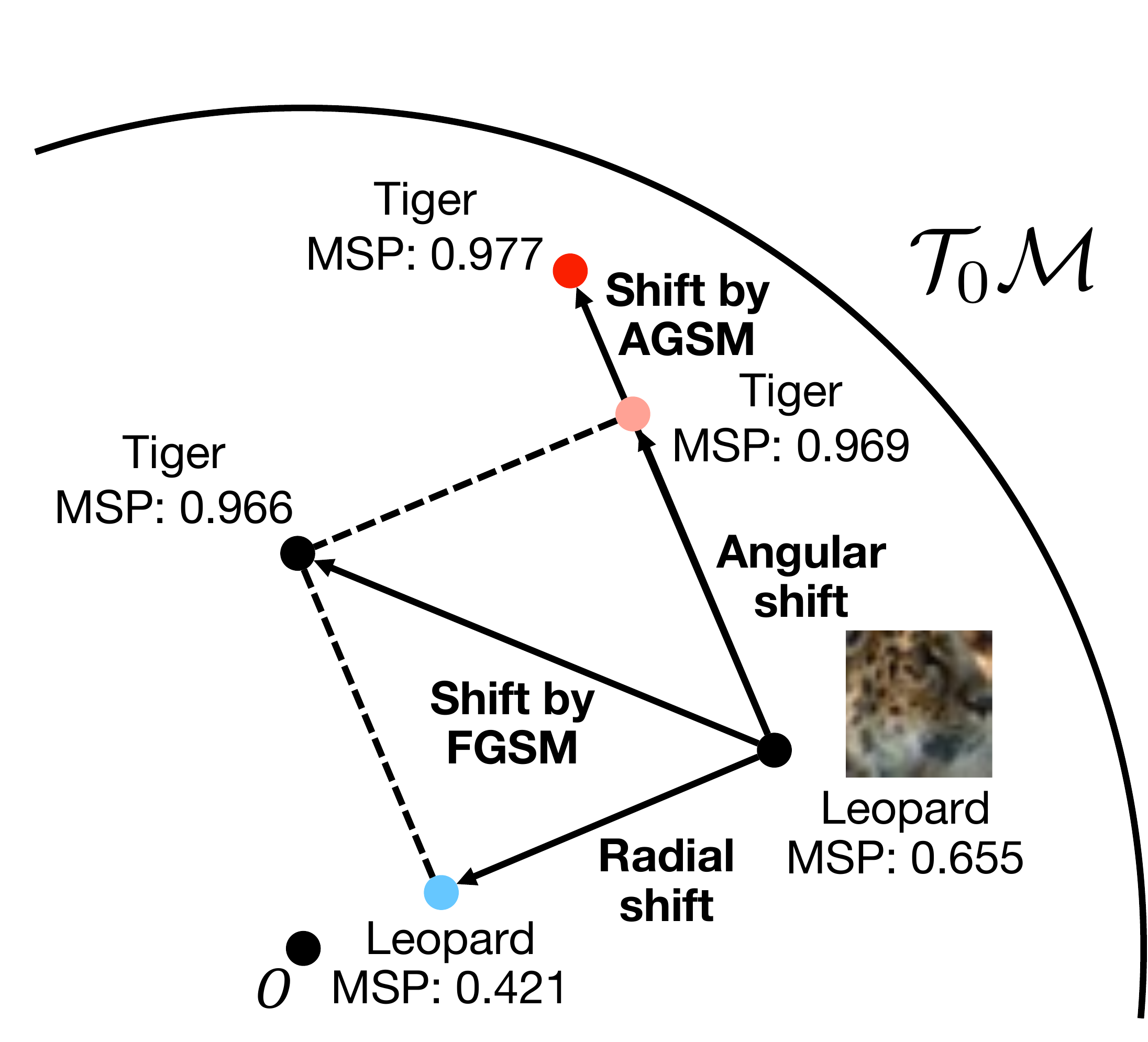} 
\caption{
\textbf{Overview of representation shifts induced by FGSM, AGSM, radial, and angular perturbations.}
We visualize how FGSM, AGSM, radial, and angular perturbations influence predictions and confidence (MSP; Maximum Softmax Probability). FGSM causes mixed, less semantic shifts, while radial perturbations reduce confidence without changing labels. AGSM amplifies angular deviation, leading to semantically meaningful misclassifications and stronger confidence drops.
}
\label{fig:fig1}
\end{figure}

However, recent advances in representation learning have demonstrated that Euclidean space is ill-suited for representing structured data, such as trees, taxonomies, or graphs, which exhibit hierarchical relationships. 
For such data, hyperbolic space provides a more natural geometric setting, offering exponential representational capacity and hierarchy-preserving structure \cite{nickel, hnn}.
This has led to the development of \textit{hyperbolic networks}, which learn and operate on latent representations embedded in hyperbolic manifolds.
These models have demonstrated strong performance not only on structured tasks such as hierarchical classification~\cite{hiercls}, knowledge graph reasoning~\cite{kngrrea}, and graph learning, but also on cross-modal retrieval tasks including text-to-image and image-to-text retrieval~\cite{desai2023meru,ramasinghe2024accept, PalSDFGM2024}. 
In these tasks, hierarchical representations are particularly important, as they capture the coarse-to-fine semantic structure that naturally arises between visual and linguistic concepts.
However, the study of adversarial robustness in hyperbolic networks remains largely unexplored~\cite{hypadv}.
Conventional adversarial attacks are geometry-agnostic, which do not account for the curvature or structure of non-Euclidean spaces.
As a result, applying them directly to hyperbolic models can lead to ineffective perturbations that do not respect the underlying geometry, resulting in representation shifts that are semantically misaligned with the structure of the manifold.

To address this limitation, we propose a novel \textit{adversarial attack that explicitly respects the geometric structure of hyperbolic space}.
Our key insight is that, in hyperbolic geometry, the loss gradient, when computed in the tangent space of a representation point, can be decomposed into a radial (depth) component and an angular (semantic) component.
The radial component alters the hierarchical level of the representation, while the angular component modulates it within the same hierarchical level, potentially aligning with semantically relevant directions in the manifold. This phenomenon is evident in Figure \ref{fig:fig1} and Table \ref{tab:tab1}. Radial shifts of the representation have negligible impact on the final prediction, whereas angular shifts account for a substantial portion of the performance degradation induced by FGSM.
Based on this decomposition, we introduce the \textit{Angular Gradient Sign Method (AGSM)}, a novel adversarial attack specifically designed for hyperbolic networks.
AGSM operates in the tangent space of hyperbolic representations and isolates only the angular component of the loss gradient, enabling perturbations that align with the semantic geometry of hyperbolic space.

This perturbation yields an adversarial example that exploits semantically sensitive directions in the hyperbolic representation space, thereby enhancing attack effectiveness without explicitly altering the hierarchical structure.
We summarize our contributions as follows:
\begin{itemize}
  \item We argue that conventional adversarial attacks may be suboptimal for hyperbolic networks, as they ignore the geometric properties of curved representation spaces and fail to exploit the structure inherent in hyperbolic embeddings.
  \item We propose AGSM (Angular Gradient Sign Method), a novel adversarial attack tailored to hyperbolic networks, which leverages radial–angular decomposition of gradients to isolate and perturb semantically sensitive directions in the hyperbolic representation space.
  \item We empirically demonstrate that AGSM outperforms conventional adversarial attacks on both hyperbolic classification and cross-modal retrieval tasks, including text-to-image and image-to-text retrieval, achieving higher fooling rates with perturbations that are more geometry-aware and effective in manipulating semantic content.
\end{itemize}

\begin{table}[t]
\centering
\begin{tabular}{l|c}
\toprule
Method                & Acc@1 (\%) \\ 
\midrule\midrule
Clean                 & 53.44      \\
FGSM                   & 19.67      \\
Radial shift       & 53.44      \\
Angular shift      & 25.56      \\
AGSM               & \textbf{13.93}      \\ 
\bottomrule
\end{tabular}
\caption{CIFAR-100 top-1 accuracy under five conditions: clean, radial shift, angular shift, FGSM, and AGSM on Poincar\'e ResNet-32. The radial shift has virtually no impact on accuracy, while the angular shift alone induces a substantial performance drop. FGSM combines both effects to further degrade accuracy, and AGSM elicits the strongest adversarial breakdown by selectively enhancing angular perturbations.
}
\label{tab:tab1}
\end{table}

\section{Related Works} 
\paragraph{Hyerpbolic Networks.}
Deep learning in hyperbolic space has demonstrated outstanding performance in encoding tree‑structured hierarchies over the recent few years~\cite{nickel, hnn, survey1, survey2, survey3}.
Several studies have shown that language and image data also exhibit hierarchical structures, and have proposed hyperbolic space as an effective solution for representing such datasets~\cite{Khrulkov_2020_CVPR, hvit, lang1, lang2}.
To facilitate more explicit learning of hierarchical structure in hyperbolic space, several studies have imposed hierarchical constraints directly on the feature representations.
As part of this research, nested geodesically convex cone~\cite{entailcone} was employed to embed directed acyclic graphs. 
\citet{visualhi} enforced hierarchical structure in image embeddings by partitioning each image into constituent parts and the overall scene for training. 
Similarly, Hi-Mapper~\cite{visrcog} preserved semantic relationships by decomposing visual scenes into individual elements and mapping them into hyperbolic space.
This trend has likewise persisted in the training of vision–language models. 
MERU~\cite{desai2023meru} jointly embeds visual and textual modalities in hyperbolic space, with the goal of encoding the language-image hierarchical relationships. 
Building on this, HyCoCLIP~\cite{PalSDFGM2024} leverages pre-trained grounding model \cite{glip1,glip2} to extract box-level image regions and corresponding text from full images and full text, more precisely structuring hierarchical relationships in hyperbolic space.
While the majority of hyperbolic networks introduce embeddings only at the penultimate layer, architectures such as Poincaré ResNet~\cite{van_Spengler_2023_ICCV}, Hyperbolic Neural Networks (HNN)~\cite{hnn,hnnpp}, HyboNet~\cite{hybo} and L-CLIP~\cite{hypco} learn their representations entirely within hyperbolic space.
While the aforementioned approaches have succeeded in embedding multiple modalities into hyperbolic space both effectively and interpretably, the robustness of these models to adversarial attacks has not been deeply explored.
\paragraph{Gradient-based Adversarial Attacks.}
Gradient-based adversarial attacks form the cornerstone of white‑box robustness evaluation in deep networks. 
The Fast Gradient Sign Method (FGSM, \citet{fgsm}) computes a single-step perturbation by taking the sign of the loss gradient with respect to the input, scaled by a budget $\varepsilon$, to maximize the model’s prediction error.
Its iterative extension, Projected Gradient Descent (PGD, \citet{pgd}), applies multiple FGSM updates of smaller magnitude, projecting the perturbed sample back onto the $\ell_p$-ball around the original input at each step, thereby yielding stronger attacks under the same perturbation constraint. 
Optimization‑based approaches such as the Carlini \& Wagner~(C\&W, \citet{carlini}) attack further refine this paradigm by framing adversarial example generation as a constrained optimization problem. 
Moreover, a variety of gradient-based adversarial attack methods have been proposed, including Jacobian-based Saliency Maps~\cite{jacob}, the Basic Iterative Method~\cite{bim}, Gradient Aligned Adversarial Subspace~\cite{subs}, Momentum Iterative FGSM~\cite{mfg}, Meta Gradient Adversarial Attack~\cite{metagr}, and Auto-Attack~\cite{auto1, auto2}.
Although these methodologies have achieved remarkable success in Euclidean space, only a handful of works have attempted to transfer these attacks into hyperbolic space.
A notable study~\cite{hypadv} applied FGM and PGD directly to synthetic hyperbolic embeddings, examining perturbation characteristics.
However, these initial efforts have largely focused on synthetic hyperbolic embeddings and input space, and have not yet considered the distinct radial and angular components in output space.
As a result, there is still an opportunity to explore how these components influence perturbation behavior and to develop attack strategies that more directly incorporate the geometric properties of hyperbolic space.
\section{Preliminaries}
We assess adversarial robustness using two representative models: Poincaré ResNet \cite{van_Spengler_2023_ICCV} and HyCoCLIP \cite{PalSDFGM2024}. 

\paragraph{Poincaré Ball Model.} 
The Poincaré ball model provides a Riemannian manifold with constant negative curvature and is widely used to model hierarchical data structures in hyperbolic neural networks. 
Notably, it serves as the geometric foundation for architectures such as Poincaré ResNet~\cite{van_Spengler_2023_ICCV}, where feature representations are embedded in hyperbolic space to capture hierarchical relations more effectively. 
Formally,  the $n$-dimensional Poincaré ball of curvature $K = -c < 0$ is defined as the open ball:
\begin{equation}
    \mathbb{B}_c^n \;=\; \bigl\{\,\bfx \in \mathbb{R}^n : \|\bfx\| < 1/c \bigr\}.
\label{eq:ball}
\end{equation}
The hyperbolic distance between two points \(u,v \in \mathbb{B}_c^n\) is given by:
\begin{equation*}
\begin{split}
  d_\mathbb{B}(\bfu,\bfv)
  &= \operatorname{arcosh}\!\Bigl(1 + 2c\,\frac{\|\bfu - \bfv\|^2}{(1 - c\|\bfu\|^2)\,(1 - c\|\bfv\|^2)}\Bigr)\\
  &= \frac{2}{\sqrt{c}}\;\tanh^{-1}\!\bigl(\sqrt{c}\,\|(-\bfu)\oplus_c \bfv\|\bigr),
\end{split}
\label{eq:poindist}
\end{equation*}
where the operation \(\oplus_c\) denotes Möbius addition in curvature \(c\).
Möbius addition generalizes the addition of the Euclidean vector to the hyperbolic space and is defined for any \(\bfx,\bfy \in \mathbb{B}_c^n\) as:
\begin{equation*}
  \bfx \oplus_c \bfy 
  = \frac{\bigl(1 + 2c\,\langle \bfx,\bfy\rangle + c\|\bfy\|^2\bigr)\,\bfx \;+\;(1 - c\|\bfx\|^2)\,\bfy}
         {1 + 2c\,\langle \bfx,\bfy\rangle + c^2\|\bfx\|^2\,\|\bfy\|^2},
\end{equation*}

\paragraph{Lorentz Model.} 
The Lorentz (hyperboloid) model offers an alternative realization of hyperbolic geometry, and is particularly useful for its numerical stability and closed-form expressions. 
This model underpins recent hyperbolic architectures such as HyCoCLIP~\cite{PalSDFGM2024}, where feature representations are embedded in Lorentzian space to effectively capture hierarchical and semantic structures.
The \(n\)-dimensional hyperbolic space of constant curvature \(K=-c<0\) is realized as the upper sheet of a two-sheeted hyperboloid in \(\mathbb{R}^{n+1}\) equipped with the Lorentzian (Minkowski) inner product:
\begin{equation*}
  \langle \bfx, \bfy \rangle_{\mathbb{L}}
  = -\,x_0\,y_0 + \sum_{i=1}^n x_i\,y_i.
\label{loin}
\end{equation*}
The Lorentz manifold is defined as:
\begin{equation*}
\begin{split}
\mathbb{L}_c^n
&= \bigl\{\,\bfx = (x_0, x_1, \dots, x_n) \in \mathbb{R}^{n+1}\;\big|\;\\
&\quad -\,x_0^2 \;+\;\sum_{i=1}^n x_i^2 \;=\; -\tfrac1c,\quad x_0 > 0
\bigr\}\,.
\end{split}
\label{lorentz}
\end{equation*}
The hyperbolic distance between two points \(\bfu,\bfv \in \mathbb{L}_c^n\) is given by:
\begin{equation}
  d_\mathbb{L}(\bfu,\bfv)
  = \frac{1}{\sqrt{c}}
    \;\operatorname{arcosh}\!\bigl(-c\,\langle \bfu, \bfv\rangle_{\mathbb{L}}\bigr).
\label{eq:hypdistlo}
\end{equation}
To perform perturbation in tangent space of Lorentz model, it is often necessary to move between the manifold and its tangent space via the exponential and logarithmic maps.
For a point \(\bfx\in\mathbb{L}_c^n\) and a tangent vector \(\bfv\in T_\bfx\mathbb{L}_c^n\) with \(\langle \bfx,\bfv\rangle_\mathbb{L}=0\), define the Lorentz norm:
\begin{equation*}
  \|\bfv\|_\mathbb{L} \;=\;\sqrt{\langle \bfv, \bfv\rangle_\mathbb{L}}\,.
\end{equation*}
Then the exponential map is:
\begin{equation}
  \exp_\bfx^c(\bfv)
  = \cosh\!\bigl(\sqrt{c}\,\|\bfv\|_\mathbb{L}\bigr)\,\bfx
    \;+\;\frac{\sinh\!\bigl(\sqrt{c}\,\|\bfv\|_\mathbb{L}\bigr)}{\sqrt{c}\,\|\bfv\|_\mathbb{L}}\;\bfv.
\label{eq:expmaplo}
\end{equation}
The corresponding logarithmic map \(\log_\bfx^c:\mathbb{L}_c^n\to T_\bfx\mathbb{L}_c^n\) for a tangent point \(\bfy\in\mathbb{L}_c^n\)
\begin{equation}
    \def\cxl{c \; \langle \bfx,\bfy \rangle _\mathbb{L}}
    \bfv = \log_\bfx(\bfy) = \frac{\cosh^{-1} (-\cxl)}{\sqrt{\left(\cxl \right)^2 - 1}} \; \text{proj}_\bfx (\bfy),
\label{logmap}
\end{equation}
where the projection onto the tangent space is defined as $\text{proj}_\bfx (\bfy) = \bfy + c \; \bfx \; \langle \bfx,\bfy \rangle_\mathbb{L}$.
These tools allow for differentiable computations and adversarial manipulations, making it a practical choice for hyperbolic deep learning.
For further details, the reader is referred to \citet{textbook}.
\section{Proposed Method}
Existing adversarial attacks are typically developed under the assumption of Euclidean geometry, making them suboptimal for models whose representations lie in curved hyperbolic spaces. 
In particular, naively applying gradient-based perturbations in hyperbolic networks may result in less effective feature shifts, as they fail to exploit the underlying hierarchical structure encoded in hyperbolic embeddings.

To address this, we propose AGSM (Angular Gradient Sign Method), a novel adversarial attack that leverages the geometric structure of hyperbolic space.
Rather than perturbing in arbitrary directions, AGSM isolates the angular component of the representation shift (i.e., the component orthogonal to the radial direction) and uses it to craft perturbations that drive semantically meaningful deviations without altering hierarchical level.
This approach is grounded in the observation that, in hyperbolic geometry, radial displacement changes hierarchical depth, whereas angular displacement induces fine-grained semantic variation within the same level. 
By explicitly targeting angular shifts, AGSM produces more effective and geometry-aware adversarial examples for hyperbolic networks.

We now formalize this approach by describing how to decompose gradients in the tangent space and apply angular perturbations in a principled manner.
\paragraph{Geometric Decomposition of FGSM Perturbations.}
To investigate how adversarial perturbations affect representations in hyperbolic space, we propose a general framework that applies to both the Poincaré and Lorentz models, two common realizations of hyperbolic geometry in neural networks.
We begin by applying the Fast Gradient Sign Method (FGSM) to generate perturbed input samples, and then analyze the resulting representation shift by decomposing it into radial and angular components in the tangent space of the corresponding manifold. 
This decomposition reflects the hierarchical and semantic structure encoded in hyperbolic embeddings and forms the basis of our geometry-aware attack method. 
Specifically, in hyperbolic space, the radial direction corresponds to changes in hierarchical depth (e.g., moving from general to specific classes), while the angular direction captures fine-grained semantic variations within the same level of the hierarchy.
By isolating the angular component, we are able to generate perturbations that exploit semantically sensitive directions in the representation space, thereby enhancing attack effectiveness without unnecessarily altering the hierarchical structure.

We begin by applying FGSM, which perturbs the input along the sign of the loss gradient:
\begin{equation*}
\mathbf{\tilde{x}}_{\mathrm{adv}}= \mathbf{x} + \varepsilon \,\mathrm{sign}\bigl(\nabla_{\mathbf{x}}\,\mathcal{L}(\theta, \mathbf{x}, y)\bigr).
\label{eq:fgsm}
\end{equation*}
Let the original and perturbed representations be defined as:
\begin{equation*}
\mathbf{h}= f(\mathbf{x}) \quad \text{and} \quad \mathbf{\tilde{h}}_{\mathrm{adv}}=f(\mathbf{\tilde{x}}_{\mathrm{adv}}),
\label{eq:encoder}
\end{equation*}
where $f(\cdot)$ denotes a hyperbolic feature encoder such as Poincaré ResNet~\cite{van_Spengler_2023_ICCV} or HyCoCLIP~\cite{PalSDFGM2024}.

We first illustrate the decomposition procedure in the case where the feature space lies in the tangent space $\mathcal{T}_0\mathbb{B}_c^n$ of the Poincaré ball  $\mathbb{B}_c^n$ (Equation~\ref{eq:ball}).
Since both $\mathbf{h}$ and $\mathbf{h}_{\mathrm{adv}}$ reside in this Euclidean tangent space, the shift in representation can be computed via:
\begin{equation*}
\Delta \mathbf{h} = \mathbf{\tilde{h}}_{\mathrm{adv}} - \mathbf{h}.
\label{eq:dh}
\end{equation*}
We then decompose $\Delta \mathbf{h}$ into radial and angular components by first computing the unit radial direction:
\begin{equation*}
\mathbf{u}_\bfh = \frac{\mathbf{h}}{\|\mathbf{h}\|_2},
\end{equation*}
and project $\Delta \mathbf{h}$ onto it to isolate the radial and angular components as:
\begin{equation*}
    \mathbf{v}_{\mathrm{rad}} = \langle \Delta \mathbf{h},\,\mathbf{u}_\mathbf{h}\rangle\,\mathbf{u}_\mathbf{h} \quad \text{and} \quad
\mathbf{v}_{\mathrm{ang}} = \Delta \mathbf{h} - \mathbf{v}_{\mathrm{rad}}.
\end{equation*}
This decomposition can be naturally extended to other hyperbolic models such as the Lorentz model. 
For networks whose output embeddings lie on the Lorentzian hyperboloid $\mathbb{L}_c^n$, we first project the hyperbolic points into the tangent space $\mathcal{T}_0\mathbb{L}^n$ using the logarithmic map (Equation~\ref{logmap}), and then perform the same radial–angular decomposition in that tangent space. 
This allows our method to generalize across different realizations of hyperbolic geometry while maintaining geometric consistency.

\paragraph{Angular-based Adversarial Perturbation.}
Building on the decomposition above, we now describe how to construct adversarial examples that explicitly maximize angular shifts in hyperbolic representation space.
Rather than perturbing the input indiscriminately in the direction of the overall gradient, as done in FGSM, we isolate the angular component $\mathbf{v}_{\mathrm{ang}}$ of the representation shift and backpropagate this direction to the input space. 
This yields an input-space gradient that selectively promotes semantic variation within the same hierarchical level.

Concretely, we compute the gradient of the inner product between the current feature representation $\mathbf{h}$ and its angular shift component $\mathbf{v}_{\mathrm{ang}}$ using the chain rule:
\begin{equation*}
\nabla_{\mathbf{x}}\bigl\langle \mathbf{h},\,\mathbf{v}_{\mathrm{ang}}\bigr\rangle
\;=\;
\left(\frac{\partial \mathbf{h}}{\partial \mathbf{x}}\right)^{\!\top}
\,\mathbf{v}_{\mathrm{ang}}.
\label{grad}
\end{equation*}
This gradient points in a direction that maximally increases the angular displacement of the representation with only negligible impact on its radial depth, thereby concentrating the perturbation on semantically meaningful variations.
We then apply a perturbation to the input in this direction, analogously to FGSM, using a normalized step:
\begin{equation*}
\mathbf{x}_{\mathrm{adv}}
= \mathbf{x}
+ \varepsilon\,\mathrm{sign}(\nabla_{\mathbf{x}}\langle \mathbf{h},\,\mathbf{v}_{\mathrm{ang}}\rangle),
\end{equation*}
where $\varepsilon$ controls the perturbation magnitude.
This method, which we term the Angular Gradient Sign Method (AGSM), results in adversarial examples that exploit semantically sensitive directions within the hyperbolic manifold. 

Compared to conventional attacks, AGSM yields more effective feature shifts while remaining aligned with the intrinsic geometry of hyperbolic space.
To formalize our method, we present the overall procedure in Algorithm~\ref{alg:agm_fgm}, which details how angular components are extracted and backpropagated to generate adversarial perturbations.

\begin{algorithm}[t!]
\caption{Angular Gradient Sign Method (AGSM)}
\label{alg:agm_fgm}
\textbf{Input}: input $\mathbf{x}$, label $y$, perturbation budget $\varepsilon$, model $f$\\
\textbf{Output}: Adversarial example $\mathbf{x}_{\mathrm{adv}}$\medskip
\begin{algorithmic}[1]
  \STATE Compute Euclidean input gradient.\\
   \quad$\mathbf{g} \gets \nabla_{\mathbf{x}}\mathcal{L}\bigl(f(\mathbf{x}),y\bigr)$
  \STATE Generate tentative perturbed input.\\
   \quad$\mathbf{\tilde{x}}_{\mathrm{adv}} \gets \mathbf{x} + \varepsilon\,\mathrm{sign}(\mathbf{g})$
  \STATE Compute feature shift.\\
   \quad$\Delta \mathbf{h} \gets f(\mathbf{\tilde{x}}_{\mathrm{adv}}) - f(\mathbf{x})$
  \STATE Get radial unit vector.\\
   \quad$\bfu \gets f(\mathbf{x}) / \lVert f(\mathbf{x})\rVert_{2}$
  \STATE Extract angular component (orthogonal to radial).\\
   \quad$\mathbf{v}_{\mathrm{ang}} \gets \Delta \mathbf{h} - \langle \Delta \mathbf{h},\bfu\rangle\,\bfu$
  \STATE Back-propagate angular shift via chain rule.\\
   \quad$\mathbf{d} \gets \bigl(\partial \mathbf{h}/\partial \mathbf{x}\bigr)^{\!\top}\,\mathbf{v}_{\mathrm{ang}}$
  \STATE Apply angular perturbation to input.\\
   \quad$\mathbf{x}_{\mathrm{adv}} \gets \mathbf{x} + \varepsilon\,\mathrm{sign}(\mathbf{d})$
  \STATE \textbf{return} $\mathbf{x}_{\mathrm{adv}}$
\end{algorithmic}
\end{algorithm}

\paragraph{Extension to Projected Gradient Descent.}
Our Angular Gradient Sign Method (AGSM) can be naturally extended into a multi-step adversarial attack by adopting the framework of Projected Gradient Descent (PGD;~\citet{pgd}).
Instead of applying a single-step angular perturbation, we iteratively maximize the angular shift at each step, followed by a projection back onto the valid perturbation set. 
This extension enables stronger adversarial examples that remain aligned with the semantic geometry of hyperbolic space while respecting perturbation constraints such as an $\ell_\infty$ budget.

At each step, the method recomputes the angular direction in feature space, backpropagates it to the input via the chain rule, and applies a normalized update, followed by a projection back into the allowed perturbation set. 
This process is repeated over multiple iterations to refine the attack and enhance its effectiveness.
The full procedure, Algorithm of Projected Angular Gradient Descent (PAGD), is summarized in supplementary material. 
\section{Experiments}
\subsection{Experimental Setup}

\begin{table}[t]
\centering
\setlength{\tabcolsep}{0.8mm}

\begin{tabular}{ccccccccc}
\toprule
Model & Dataset & $\varepsilon$ & & FGSM & AGSM & PGD & PAGD & Clean \\
\midrule \midrule
\multirow{9}{*}{{\shortstack{PRN\\20}}} 
& \multirow{3}{*}{C-10}
& $\varepsilon_1$  && 56.59 & \textbf{47.63} & 31.83 & \textbf{22.42} & \multirow{3}{*}{84.76} \\
& & $\varepsilon_2$  && 54.43 & \textbf{44.53} & 21.40 & \textbf{14.25} & \\
& & $\varepsilon_3$  && 49.63  &\textbf{ 36.69}  & 8.86  & \textbf{8.35}  & \\
\cmidrule{2-9}
& \multirow{3}{*}{C-100}
& $\varepsilon_1$  && 24.67 & \textbf{20.02 }& 13.29 & \textbf{11.05} & \multirow{3}{*}{49.63} \\
& & $\varepsilon_2$  && 22.62 & \textbf{17.66} & 11.68 & \textbf{9.28}  & \\
& & $\varepsilon_3$  && 17.78 & \textbf{12.19}  & 9.43  & \textbf{9.43}  & \\
\cmidrule{2-9}
& \multirow{3}{*}{TIN}
& $\varepsilon_1$  && 11.73 & \textbf{8.90 } & 7.31  & \textbf{5.93}  & \multirow{3}{*}{30.48} \\
& & $\varepsilon_2$  && 10.50 & \textbf{7.78}  & 6.62  & \textbf{5.49 } & \\
& & $\varepsilon_3$  && 7.44  & \textbf{5.43 } & 5.66  & \textbf{4.63}  & \\
\midrule
\multirow{9}{*}{{\shortstack{PRN\\32}}} 
& \multirow{3}{*}{C-10}
& $\varepsilon_1$  && 60.68 & \textbf{51.09} & 28.96 & \textbf{18.69} & \multirow{3}{*}{86.21} \\
& & $\varepsilon_2$  && 59.10 & \textbf{48.05} & 18.43 & \textbf{11.44} & \\
& & $\varepsilon_3$  && 54.19 & \textbf{41.56}  & 8.05  & \textbf{7.77}  & \\
\cmidrule{2-9}
& \multirow{3}{*}{C-100}
& $\varepsilon_1$  && 26.36 & \textbf{21.05} & 12.71 & \textbf{10.44} & \multirow{3}{*}{53.44} \\
& & $\varepsilon_2$  && 24.61 & \textbf{18.74} & 11.18 & \textbf{9.19}  & \\
& & $\varepsilon_3$  && 19.67 & \textbf{13.93}  & 9.24  &\textbf{ 7.86}  & \\
\cmidrule{2-9}
& \multirow{3}{*}{TIN}
& $\varepsilon_1$  && 11.90 & \textbf{9.56}  & 7.03  & \textbf{5.74}  & \multirow{3}{*}{30.46} \\
& & $\varepsilon_2$  && 10.71 & \textbf{8.23}  & 6.61    & \textbf{5.49}    & \\
& & $\varepsilon_3$  && 8.02  & \textbf{5.57}  & 5.69    & \textbf{5.00}    & \\
\bottomrule
\end{tabular}
\caption{Robust accuracy (\%) of {Poincaré ResNet-20} and {ResNet-32} on CIFAR-10, CIFAR-100, and Tiny ImageNet under $\ell_\infty$ attacks with $\varepsilon \in \{2.4/255, 3.2/255, 8.0/255\}$. For each attack type (FGSM and PGD), the lower accuracy (indicating a stronger attack) is highlighted in \textbf{bold}.}
\label{tab:main_cifar}
\end{table}

\paragraph{Datasets.}
To evaluate our method on standard image classification benchmarks, we use CIFAR‑10, CIFAR‑100 \cite{cifar} and Tiny ImageNet \cite{tiny}, covering a range of object categories and difficulty levels. For image‐to‐text (I2T) and text‐to‐image (T2I) retrieval experiments, we conduct evaluations on the MS COCO dataset \cite{coco} and the Flickr30K dataset \cite{flickr}, which provide paired image–caption annotations suitable for cross‐modal retrieval tasks.  
\paragraph{Models.} For the image classification experiments, we employ Poincaré ResNet‐20 and Poincaré ResNet‐32 architectures, both of which we trained using the exact hyperparameter settings and training protocol specified in the original \textbf{Poincaré ResNet}~\cite{van_Spengler_2023_ICCV}. For cross‐modal retrieval (I2T and T2I), we utilize the \textbf{HyCoCLIP} framework \cite{PalSDFGM2024} with Vision Transformer backbones (ViT‐S and ViT/16), leveraging the pretrained weights officially released by the HyCoCLIP authors.  

\begin{table*}[t]
  \centering
  \setlength{\tabcolsep}{1mm}
    \begin{tabular}{ccc
                    ccccc
                    ccccc}
      \toprule
      \multirow{2}{*}{{Model}}     & \multirow{2}{*}{{Dataset}}  & \multirow{2}{*}{{$\varepsilon$}}
        & \multicolumn{5}{c}{T2I R@5}
        & \multicolumn{5}{c}{T2I R@10} \\
      \cmidrule(lr){4-8} \cmidrule(lr){9-13}
               &          &
        & FGSM   & AGSM & PGD   & PAGD & Clean
        & FGSM   & AGSM & PGD   & PAGD & Clean \\
      \midrule\midrule
      \multirow{4}{*}{ViT‑S/16}
      & \multirow{2}{*}{COCO}
              & 3.2/255 & 11.20 & \textbf{8.20} &  3.00 &  \textbf{2.60} & \multirow{2}{*}{55.10}
                 & 16.50 & \textbf{12.70} & 5.10 & \textbf{4.40} & \multirow{2}{*}{66.60}\\
      &        & 8.0/255 & 7.60  & \textbf{4.80} &  1.50 &  \textbf{1.00}
                  &       & 11.60 & \textbf{7.60}  & 2.70 & \textbf{1.90} & \\
      \cmidrule(lr){2-13}
      & \multirow{2}{*}{Flickr30K}
              & 3.2/255 & 19.70 & \textbf{15.40} & 8.40 & \textbf{7.30} & \multirow{2}{*}{81.50}
                 & 27.30 & \textbf{22.50} & 13.50 & \textbf{12.10} & \multirow{2}{*}{88.20}\\
      &        & 8.0/255 & 12.90 & \textbf{9.10} & 4.40 & \textbf{4.00}
                  &       & 19.10 & \textbf{14.30} & 7.50 & \textbf{7.00} & \\
      \midrule
      \multirow{4}{*}{ViT‑B/16}
      & \multirow{2}{*}{COCO}
              & 3.2/255 & 15.90 & \textbf{12.60} & 4.50 & \textbf{4.00} & \multirow{2}{*}{58.40}
                 & 22.70 & \textbf{18.80} & 7.30 & \textbf{6.40} & \multirow{2}{*}{69.30}\\
      &        & 8.0/255 & 10.80 & \textbf{7.60} & 2.20 & \textbf{1.80}
                  &       & 16.20 & \textbf{11.90} & 3.80 & \textbf{3.10} & \\
      \cmidrule(lr){2-13}
      & \multirow{2}{*}{Flickr30K}
              & 3.2/255 & 27.20 & \textbf{24.90} & 10.90 & \textbf{9.80} & \multirow{2}{*}{84.90}
                 & 36.30 & \textbf{33.20} & 17.70 & \textbf{16.00} & \multirow{2}{*}{90.30}\\
      &        & 8.0/255 & 18.60 & \textbf{14.10} & 5.30 & \textbf{5.00}
                  &       & 25.10 & \textbf{21.00} & 9.50 & \textbf{8.40} & \\
      \bottomrule
    \end{tabular}
  \caption{Performance of the Text-to-Image (T2I) task at Recall@5 and Recall@10 under adversarial attacks (FGSM, AGSM, PGD, PAGD) on COCO and Flickr30K using ViT-S/16 and ViT-B/16. For each attack type (FGSM and PGD), the lower accuracy (indicating a stronger attack) is highlighted in \textbf{bold}.}
  \label{tab:main_clip_t2i}
\end{table*}

\begin{table*}[t]
  \centering
  
  \setlength{\tabcolsep}{1mm}
  
    \begin{tabular}{ccc
                    ccccc
                    ccccc}
      \toprule
      \multirow{2}{*}{{Model}}     & \multirow{2}{*}{{Dataset}}  & \multirow{2}{*}{{$\varepsilon$}}
        & \multicolumn{5}{c}{I2T R@5}
        & \multicolumn{5}{c}{I2T R@10} \\
      \cmidrule(lr){4-8} \cmidrule(lr){9-13}
               &          &
        & FGSM   & AGSM & PGD   & PAGD & Clean
        & FGSM   & AGSM & PGD   & PAGD & Clean \\
      \midrule\midrule
      \multirow{4}{*}{ViT‑S/16}
      & \multirow{2}{*}{COCO}
              & 3.2/255 & 13.20 & \textbf{9.00} & 4.10 & \textbf{3.10} & \multirow{2}{*}{69.50}
                 & 18.50 & \textbf{13.60} & 6.10 & \textbf{4.70} & \multirow{2}{*}{79.50} \\
      &        & 8.0/255 & 7.30 & \textbf{4.20} & 1.90 & \textbf{1.40} &       
                  & 11.10 & \textbf{6.80} & 3.10 & \textbf{2.30} &       \\
      \cmidrule(lr){2-13}
      & \multirow{2}{*}{Flickr30K}
              & 3.2/255 & 20.10 & \textbf{15.30} & 8.10 & \textbf{6.90} & \multirow{2}{*}{89.10}
                 & 27.10 & \textbf{21.70} & 13.30 & \textbf{9.90} & \multirow{2}{*}{93.90} \\
      &        & 8.0/255 & 11.60 & \textbf{8.50} & 4.20 & \textbf{3.90} &       
                  & 16.70 & \textbf{12.20} & 7.70 & \textbf{7.20} &       \\
      \midrule
      \multirow{4}{*}{ViT‑B/16}
      & \multirow{2}{*}{COCO}
              & 3.2/255 & 20.10 & \textbf{15.00} & 5.40 & \textbf{4.60} & \multirow{2}{*}{72.00}
                 & 26.90 & \textbf{21.60} & 8.30 & \textbf{7.50} & \multirow{2}{*}{82.00} \\
      &        & 8.0/255 & 11.10 & \textbf{7.40} & 2.40 & \textbf{1.90} &       
                  & 15.90 & \textbf{10.90} & 3.80 & \textbf{3.20} &       \\
      \cmidrule(lr){2-13}
      & \multirow{2}{*}{Flickr30K}
              & 3.2/255 & 29.50 & \textbf{26.60} & 11.70 & \textbf{10.70} & \multirow{2}{*}{92.60}
                 & 38.40 & \textbf{35.00} & 16.90 & \textbf{16.20} & \multirow{2}{*}{95.40} \\
      &        & 8.0/255 & 18.10 & \textbf{13.70} & 6.10 & \textbf{4.40} &       
                  & 24.60 & \textbf{19.70} & 9.00 & \textbf{7.30} &       \\
      \bottomrule
    \end{tabular}
  \caption{Performance of the Image-to-Text (I2T) task at Recall@5 and Recall@10 under adversarial attacks (FGSM, AGSM, PGD, PAGD) on COCO and Flickr30K using ViT-S/16 and ViT-B/16. }
  \label{tab:main_clip_i2t}
\end{table*}

\subsection{Results on Classification and Retrieval Tasks}
\paragraph{Poincaré ResNet Robustness (Table \ref{tab:main_cifar}).}

Across both ResNet-20 and ResNet‐32 on CIFAR-10, Angle-only FGSM (AGSM) consistently inflicts an extra 9–11\% drop in robust accuracy over standard FGSM, while PAGD compounds PGD’s effect by roughly the same amount (around 9-10\%). For instance, at $\varepsilon=8.0/255$ on ResNet-32, AGSM lowers clean accuracy \textbf{12.63\%} more than FGSM, and PAGD further lowers clean accuracy \textbf{10.27\%} more than PGD at $\varepsilon=2.4/255$. ResNet-20 shows a similar pattern, with AGSM undercutting FGSM by about \textbf{13\%} and PAGD lowering PGD by over \textbf{9\%} in its strongest case.

On CIFAR-100 and Tiny ImageNet, AGSM still outperforms its one‐step counterpart by about 5–6\%, and PAGD delivers an additional 1–2\% degradation beyond PGD. These results confirm that angular‐maximizing perturbations more effectively exploit hyperbolic geometry than conventional gradient‐based methods.
\paragraph{HyCoCLIP Retrieval Robustness (Table \ref{tab:main_clip_t2i}, \ref{tab:main_clip_i2t}).}
Across both COCO and Flickr30K, and for both ViT-S/16 and ViT-B/16 backbones, our Angle-only FGSM (AGSM) consistently deepens the drop in recall by roughly 2–5\% compared to standard FGSM, while its multi-step counterpart PAGD yields an additional 0.5–1\% reduction over PGD.

For example, on COCO with ViT-S/16 at $\varepsilon=3.2/255$, AGSM adds a further 3.0\% of degradation in T2I R@5 beyond FGSM’s already severe drop, and PAGD compounds PGD’s effect by another 0.8\%. With the larger ViT-B/16, AGSM’s advantage reaches up to about a 5\% extra drop in I2T R@5, and PAGD still provides nearly a 1\% degradation over PGD. These largest observed gains underline that angular-maximising perturbations more effectively disrupt cross-modal retrieval in hyperbolic embedding spaces.
\paragraph{Summary.}
Across both Poincaré ResNet and HyCoCLIP backbones, standard one-step (FGSM) and multi-step (PGD) attacks already degrade performance, but their angular maximizing counterparts, AGSM and PAGD, consistently inflict an additional drop in accuracy or recall. This highlights the critical role of angular movement in breaking hierarchical representations.
\subsection{Analysis on Perturbed Sample and Representation}
\begin{figure}[t]
\centering
\includegraphics[width=\columnwidth]{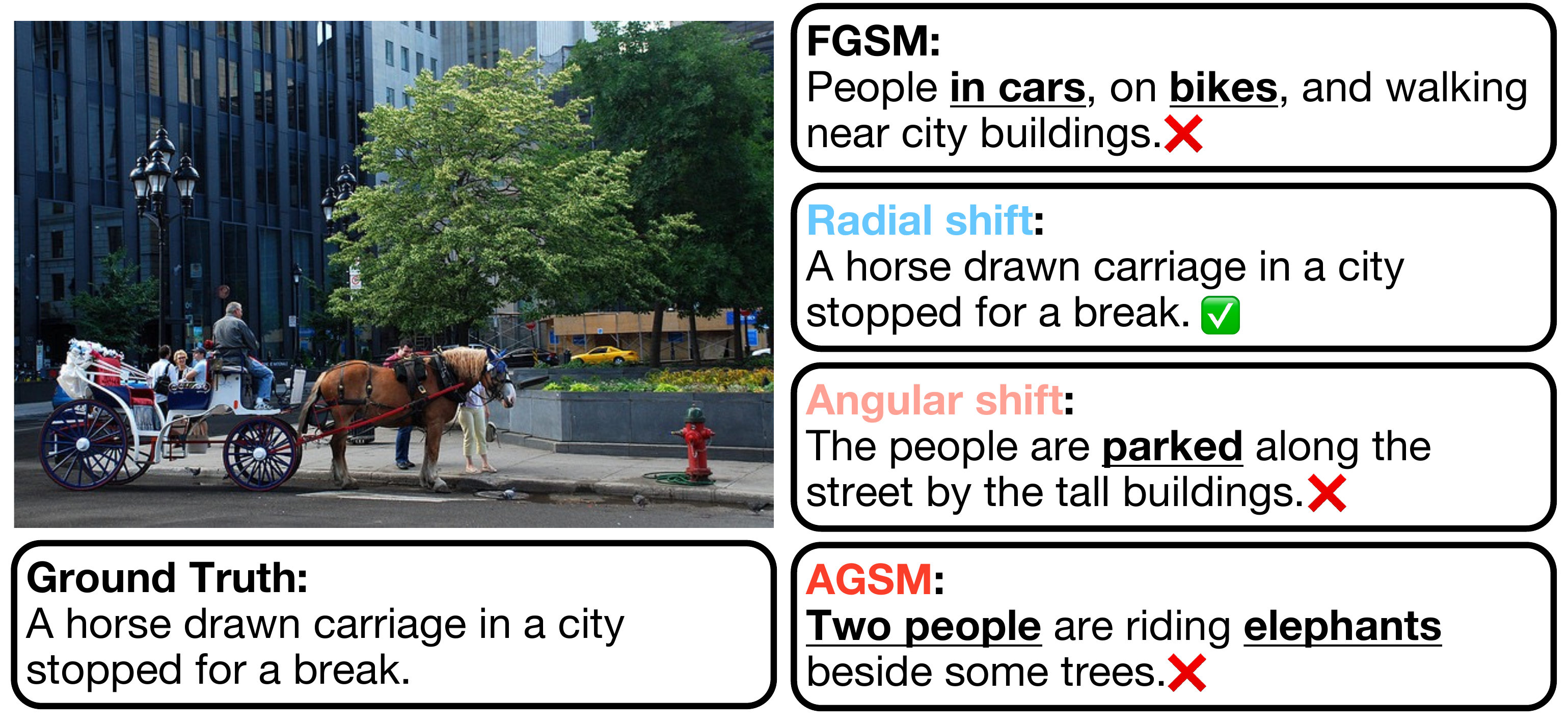} 
\caption{Qualitative comparison of Image-to-Text retrieval under FGSM, radial shift, angular shift, and AGSM. While the radial shift preserves the correct caption, FGSM and the standard angular shift generate semantically incorrect outputs, and AGSM yields the most misaligned caption.}
\label{fig:fig2}
\end{figure}

\paragraph{Distance Between Hyperbolic Embeddings.}  
Table \ref{tab:hyperbolic-feature-distances} reports the average hyperbolic distance in the Lorentz model (Equation \ref{eq:hypdistlo}) between original and perturbed feature vectors, after mapping them to the hyperbolic manifold via the exponential map (Equation \ref{eq:expmaplo}). Features were produced by HyCoCLIP under FGSM and AGSM at $\varepsilon\in\{3.2/255,8.0/255\}$. On both dataset, AGSM increases the mean geodesic at both $\varepsilon$ values, indicating that angular‐maximizing updates push representations farther along hyperbolic geodesics than standard gradient‐sign perturbations. Figure \ref{fig:fig2} qualitatively compares retrieval outputs under different perturbations. The radial shift preserves the correct caption, FGSM and the standard angular shift yield semantically incorrect sentences, and AGSM produces the most semantically misaligned caption. 
\begin{table}[t]
  \centering
  \begin{tabular}{c c c c}
    \toprule
    Dataset   & $\varepsilon$ & Clean vs FGSM & Clean vs AGSM \\
    \midrule\midrule
    COCO  & 3.2/255    & 0.3058      & \textbf{0.3639}      \\
              & 8.0/255    & 0.3883      & \textbf{0.4457}      \\
    \cmidrule(lr){1-4}
    Flickr30K & 3.2/255    & 0.3119      & \textbf{0.3675}      \\
              & 8.0/255    & 0.3875      & \textbf{0.4400}      \\
    \bottomrule
  \end{tabular}
  \caption{Hyperbolic feature distances (in $\mathbb{L}_c^n$) between the original and perturbed samples for HyCoCLIP (ViT/b/16) under FGSM and AGSM.}
  \label{tab:hyperbolic-feature-distances}
\end{table}
\paragraph{Confidence Drop.}  
\begin{table}[t]
  \centering
  \begin{tabular}{c c c c}
    \toprule
    Dataset   & $\varepsilon$ & FSGM   & AGSM    \\
    \midrule\midrule
    \multirow{2}{*}{CIFAR‑10}
      & 3.2/255 & 0.3870 & \textbf{0.4860} \\
      & 8.0/255 & 0.4364 & \textbf{0.5597} \\
    \cmidrule(lr){1-4}
    \multirow{2}{*}{CIFAR‑100}
      & 3.2/255 & 0.4242 & \textbf{0.4628} \\
      & 8.0/255 & 0.4566 & \textbf{0.4935} \\
    \bottomrule
  \end{tabular}
  \caption{Under FGSM and AGSM, drop in MSP of the model on initially predicted label.}
  \label{tab:confidence-drop}
\end{table}

Table \ref{tab:confidence-drop} shows that AGSM consistently produces larger confidence reductions than FGSM across both CIFAR-10 and CIFAR-100, and at both moderate and high perturbation levels. In particular, the gap between FGSM and AGSM widens as $\varepsilon$ increases, indicating that emphasizing the angular shift becomes even more destructive under stronger attacks. Qualitatively, this demonstrates that angular-focused perturbations more effectively undermine the model’s predictive certainty than conventional gradient-sign methods. 
\paragraph{Summary.} Table \ref{tab:hyperbolic-feature-distances} and Table \ref{tab:confidence-drop} together underscore the central contribution of our approach: by isolating and maximizing the angular component of the gradient, AGSM not only drives feature vectors to traverse significantly farther along hyperbolic geodesics but also precipitates a more severe collapse of the model’s predictive confidence compared to conventional gradient-sign attacks. This dual effect confirms that angular-maximizing perturbations provide a principled mechanism to undermine both representational integrity and output certainty in hierarchical models.
\begin{table}[t!]
  \centering
  \setlength{\tabcolsep}{0.8mm}
  \begin{tabular}{c c c c c c c}
    \toprule
    Dataset   & $\varepsilon$ & FGSM    & AGSM & PGD    & PAGD & Clean \\
    \midrule\midrule
    \multirow{2}{*}{C‑10}
      & 3.2/255  & 58.20  & \textbf{49.38}                & 32.79  & \textbf{25.47}                & \multirow{2}{*}{84.76} \\
      & 8.0/255  & 51.74  & \textbf{40.16}                & 10.06  & \textbf{8.62}                &                         \\
    \midrule
    \multirow{2}{*}{C‑100}
      & 3.2/255  & 24.97  & \textbf{20.96}                & 13.34  & \textbf{10.84}                & \multirow{2}{*}{49.63} \\
      & 8.0/255  & 19.43  & \textbf{14.31}                & 9.46  & \textbf{7.74}                &                         \\
    \bottomrule
  \end{tabular}
  \caption{Top‑1 accuracy (\%) of Poincaré ResNet‑20 under $\ell_2$‑constrained Fast Gradient Sign Method (FGSM) and Projected Gradient Descent (PGD), together with their angular‐only variants, on CIFAR‑10 and CIFAR‑100. “Original” denotes clean accuracy.}
  \label{tab:l2-attacks-poincare20}
\end{table}

\subsection{Ablation Study}
Under an $\ell_{2}$‐constraint (Table \ref{tab:l2-attacks-poincare20}), results demonstrate that isolating and maximizing the angular component delivers powerful attacks that are largely agnostic to the choice of norm. Whether measured in $\ell_\infty$ or $\ell_2$, AGSM consistently exploits angular vulnerabilities in hyperbolic embeddings more effectively than FSGM.  
\begin{table}[t]
\centering
\setlength{\tabcolsep}{1mm}
\begin{tabular}{ccccc}
\toprule
\multirow{2}{*}{{Dataset}} & \multirow{2}{*}{{Attack}} & \multicolumn{3}{c}{{Training Data}} \\
\cmidrule(lr){3-5}
& & {Clean} & {FGSM-Aug} & {AGSM-Aug} \\
\midrule\midrule
\multirow{3}{*}{CIFAR-10}
& Clean         & \textbf{84.76} & 81.65 & 82.31 \\
& FGSM          & 8.67  & \textbf{56.58} & 55.08 \\
& AGSM          & 8.30  & \textbf{52.46} & 51.07 \\
\midrule
\multirow{3}{*}{CIFAR-100}
& Clean         & \textbf{49.63} & 47.81 & 46.96 \\
& FGSM          & 9.61  & 26.45 & \textbf{27.99} \\
& AGSM          & 7.91  & 23.61 & \textbf{25.66} \\
\bottomrule
\end{tabular}
\caption{Adversarial Training Results. Top-1 accuracy (\%) of Poincaré ResNet-20 trained on clean, FGSM-augmented, and AGSM-augmented datasets, evaluated under clean, FGSM, and AGSM attacks on CIFAR-10 and CIFAR-100.}
\label{tab:adv_training_combined}
\end{table}

\section{Conclusion and Limitation}
In this work, we introduced the Angular Gradient Sign Method (AGSM) and its multi‐step extension PAGD to craft adversarial perturbations that explicitly maximize angular shifts in hyperbolic embedding spaces. Through extensive experiments on Poincaré ResNet for image classification and HyCoCLIP for cross‐modal retrieval, we demonstrated that angular‐focused attacks consistently outperform standard FGSM and PGD baselines. Our ablation studies further revealed that the angular component alone drives the majority of the adversarial effect, and that these attacks remain effective under both $\ell_{\infty}$ and $\ell_{2}$ norm constraints. 

However, training with AGSM‐perturbed examples yields only modest gains in robustness and incurs a trade‐off in clean accuracy relative to FGSM augmentation. On CIFAR-100, AGSM augmentation improves robustness specifically against angular perturbations but at the cost of a larger drop in standard accuracy. These results suggest that naively incorporating adversarial examples perturbed by AGSM does not uniformly strengthen hyperbolic models and may incur dataset‐dependent trade‐offs.
Taken together, our findings underscore the pivotal role of angular misalignment in hyperbolic vulnerability and point to the need for geometry‐aware defense strategies that explicitly accommodate the curved, hierarchical structure of hyperbolic embeddings.

\section*{Acknowledgements}
This work was supported by the National Research Foundation of Korea (NRF) grant funded by the Korea government (MSIT) (No. RS-2024-00345809, ``Research on AI Robustness Against Distribution Shift in Real-World Scenarios''), the Institute of Information \& Communications Technology Planning \& Evaluation (IITP) grant funded by the Korea government (MSIT) (No. RS-2025-02263754, ``Human-Centric Embodied AI Agents with Autonomous Decision-Making''),and the Korea Health Technology R\&D Project through the Korea Health Industry Development Institute (KHIDI), funded by the Ministry of Health \& Welfare, Republic of Korea (No. RS-2025-02307233).

\bibliography{aaai2026}

\clearpage
\appendix
\section{Algorithm of PAGD}

\begin{algorithm} [H]
\caption{Projected Angular Gradient Descent (PAGD)}
\label{alg:agm_pgd}
  \textbf{Input}: input $\mathbf{x}$, label $y$, perturbation budget $\varepsilon$, step size $\alpha$, iterations $T$, model $f$\\
  \textbf{Output}: Adversarial example $\mathbf{x}_{T}$\medskip
\begin{algorithmic}[1]
  \STATE $\mathbf{x}_0 \gets \mathbf{x}$
  \FOR{$t = 0,1,\dots,T-1$}
    \STATE Compute representation shift. \\ 
    \quad $\Delta \mathbf{h}_t \gets f(\mathbf{x}_t) - f(\mathbf{x}_{t-1})$
    \STATE Compute radial unit vector.\\ 
    \quad$\mathbf{u}_{h,t} \gets \dfrac{f(\mathbf{x}_t)}{\|f(\mathbf{x}_t)\|_2}$
    \STATE Compute angular component.\\ 
    \quad $\mathbf{v}_{\mathrm{ang},t} \gets \Delta \mathbf{h}_t - \langle \Delta \mathbf{h}_t,\,\mathbf{u}_{h,t}\rangle\,\mathbf{u}_{h,t}$
    \STATE Backpropagate to input.\\
    \quad $\mathbf{d}_t \;\gets\;\bigl(\partial f(\mathbf{x}_t)/\partial {\mathbf{x}_t}\bigr)^{\!\top}\,\mathbf{v}_{\mathrm{ang},t}$
    \STATE Angular update.\\
    \quad $\tilde{\mathbf{x}}_{t+1} \gets \mathbf{x}_t + \alpha\,\mathrm{sign}(\mathbf{d}_t)$
    \STATE Project.\\
    \quad $\mathbf{x}_{t+1} \gets \Pi_{\|\cdot - \mathbf{x}\|\le\varepsilon}\bigl(\tilde{\mathbf{x}}_{t+1}\bigr)$
  \ENDFOR
  \STATE \textbf{return} $\mathbf{x}_T$
\end{algorithmic}
\end{algorithm}
\section{Implementation Details}
To evaluate adversarial robustness in image classification models, we followed the experimental setup from prior work \cite{van_Spengler_2023_ICCV} and used Poincaré ResNet-20 and ResNet-32 pretrained on CIFAR-10, CIFAR-100, and Tiny ImageNet. For Tiny ImageNet, all images were resized to 32×32 during training for consistency. For retrieval tasks, we assessed the adversarial robustness of HyCoCLIP using ViT-B and ViT-S backbones, initialized with the pretrained weights provided by the authors of HyCoCLIP \cite{PalSDFGM2024}.

Unless otherwise specified in each table, the perturbation budget $\varepsilon$ was fixed at 8.0/255. When applying PGD to image classification models, we used $T = 20$ iterations with a step size of $\alpha = \varepsilon/4$. In the retrieval setting, PGD was applied with $T = 20$ and a smaller step size of $\alpha = \varepsilon/10$.

All experiments were conducted using PyTorch 2.6.0 and a single NVIDIA GeForce RTX 4090 (24GB).

\end{document}